\documentclass[runningheads]{llncs}

\usepackage{eccv}

\usepackage{eccvabbrv}

\usepackage{graphicx}
\usepackage{booktabs}
\usepackage{multirow}
\usepackage[accsupp]{axessibility}  

\usepackage{hyperref}

\usepackage{orcidlink}

\begin{document}

\title{A vision-based framework for human behavior understanding in industrial assembly lines} 

\titlerunning{Visual human behavior understanding in industrial assembly lines}

\author{Konstantinos Papoutsakis\inst{1}\orcidlink{0000--0002--2467--8727} \and
Nikolaos Bakalos\inst{2,3}\orcidlink{0000--0002--3106--4758} \and
Konstantinos Fragkoulis \inst{2}\orcidlink{0009--0006--8833--1411} \and
Athena Zacharia \inst{2,3}\orcidlink{0009--0009--9295--471X} 
\and
Georgia Kapetadimitri\inst{2} \and
Maria Pateraki \inst{2,3}\orcidlink{0000--0002--8943--4598} }

\authorrunning{K.~Papoutsakis et al.}

\institute{Institute of Computer Science, Foundation for Research and Technology - Hellas, Greece \email{papoutsa@ics.forth.gr} \and
School of Rural Surveying and Geoinformatics Engineering, National Technical University of Athens, Greece \and
Institute of Communication and Computer Systems, National Technical University of
Athens, Greece\\
\email{\{bakalosnik,kfragoulis,azacharia, georgiakpd, mpateraki\}@mail.ntua.gr}}

\maketitle

\begin{abstract}
This paper introduces a vision-based framework for capturing and understanding human behavior in industrial assembly lines, focusing on car door manufacturing. The framework leverages advanced computer vision techniques to estimate workers' locations and 3D poses and analyze work postures, actions, and task progress. A key contribution is the introduction of the CarDA dataset, which contains domain-relevant assembly actions captured in a realistic setting to support the analysis of the framework for human pose and action analysis. The dataset comprises time-synchronized multi-camera RGB-D videos, motion capture data recorded in a real car manufacturing environment, and annotations for EAWS-based ergonomic risk scores and assembly activities.  
Experimental results demonstrate the effectiveness of the proposed approach in classifying worker postures and robust performance in monitoring assembly task progress. 
\end{abstract}

\section{Introduction}
\label{sec:intro}

In industrial manufacturing, the continuous advancement of automation technologies necessitates a comprehensive understanding of human behavior within assembly lines to enhance productivity and safety. This paper introduces a vision-based framework designed to analyze and interpret human behaviors specifically in the context of industrial assembly lines. By leveraging advanced computer vision techniques, this framework aims to monitor and assess physical ergonomic, and operational aspects of human activities during the assembly process.
The framework employs state-of-the-art methods for human pose estimation, ergonomic postural evaluation, and human action monitoring, providing a robust solution to the challenges of human dynamics, severe long-term human body occlusions, and understanding of complex human assembly actions. The integration of these methods facilitates the real-time assessment of workers' postures and actions, enabling the identification of potential ergonomic risks and inefficiencies in the workflow. An important contribution of this research is the development and deployment of the CarDA dataset—a comprehensive multi-modal dataset comprising time-synchronized RGB-D videos and motion capture data recorded in a real-world car manufacturing environment providing in a single dataset ground truth data for 3D human poses, ergonomic risk scores based on the European Assessment Work Sheet (EAWS) posture grid, temporal segmentation, and the classification of assembly actions demonstrated by real workers. Most available datasets published so far deal with daily activities~\cite{Liu_2019_NTURGBD120,carreira2017quo, tenorth09dataset} and very few have been published on manufacturing actions, usually adopting an egocentric perspective focusing on finer movements of hands~\cite{schoonbeek2024industreal,ragusa2024enigma, sener2022assembly101,cicirelli2022ha4m,10160633,9897369}. At the same time, our framework requires camera perspectives overlooking the workstation area where the activity is performed to capture large body movements and human-car door interactions during the assembly task cycle. Furthermore, available datasets are strongly domain-related to specific assembly activities, not transferable to the car door assembly task. 
This dataset is a critical resource for training and evaluating the proposed vision-based framework, ensuring its applicability and effectiveness in real industrial settings. The dataset will be available freely available online~\footnote{~\url{https://zenodo.org/uploads/13370888}}.

\begin{figure}[t]
    \centering
    \includegraphics[width=\textwidth]{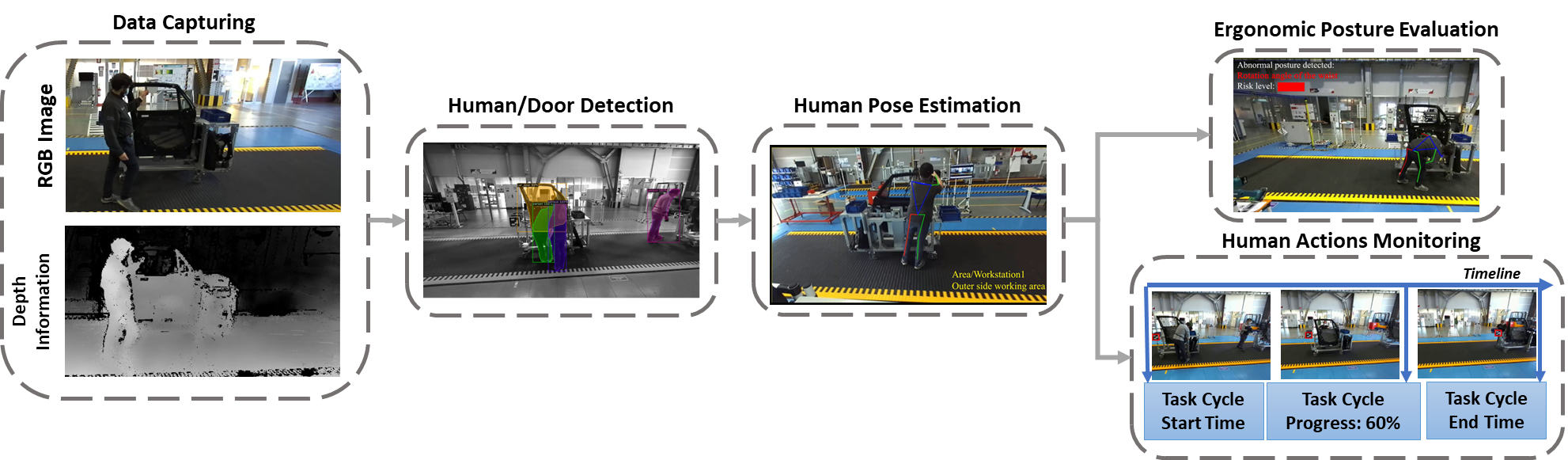}
    \caption{Workflow of the human behavior understanding framework.}
    \label{fig:hbu_workflow}
\end{figure}

\section{Related work}
\subsection{Human pose estimation}
\label{sec:sota_detection}
Human pose estimation (HPE) in 2D or 3D, in images or from videos is a challenging yet fundamental Computer Vision task due to depth ambiguity and self-occlusion and the fact that it is an inverse, highly ill-posed problem with multiple feasible solutions. 
In addition, the rather time-consuming process of collecting high-quality and fine-grained annotations (labeled/ground truth data) for the human body leads to limited access to training data. 
In particular, annotating 3D data related to human motion requires dedicated, costly motion capture systems, which are not easy to obtain and install to capture human motion and activities in various contexts and environmental conditions.
The surveys of~\cite{POPPE2010976,MOESLUND200690,liu2022recent,ZhangSensors2019,Lei2019ASO} reviewed the human motion analysis in many aspects (e.g., detection and tracking, pose estimation, recognition). They described the relation between human pose estimation and other related tasks.
More recent surveys mainly focused on subdomains, such as RGB-D-based action recognition \cite{wang2018rgb}, 3D human pose estimation \cite{shin2023wham, tripathi2023ipman,wang23refit,gong2023diffpose}, model-based human pose estimation \cite{Qammaz2023b}, body parts-based human pose \cite{Qammaz2020,zheng2023deep}, and monocular- based human pose estimation \cite{pavlakos2018learning}. Despite the fairly accurate performance of state-of-the-art algorithms for 3D human pose estimation in controlled or semi-controlled settings, coping with complex, realistic scenarios exposes the limits of these algorithms, particularly their effectiveness in handling occlusions. The latter is a key aspect of our approach (sec.~\ref{sec:human_cardoor}) aiming to derive valid detections in the presence of occlusions by utilizing synchronized camera data.

\subsection{Vision-based ergonomic posture evaluation}
\label{sec:sota_risk}
Several screening tools for physical ergonomic analysis are commonly used in industry, based on different evaluation protocols, sets of postures and other related human activities such as grasping and lifting, such as the Rapid Entire Body Assessment (REBA)~\cite{mcatamney2004rapid}, the European Assembly Worksheet~\cite{schaub2013european} (EAWS), the OCRA checklist~\cite{occhipinti1998ocra}, the MURI risk analysis~\cite{womack2006lean} and the Ovako Working Posture Analyzing System (OWAS)~\cite{karhu1977correcting}. 
Based on these methods, various efficient vision-based approaches have been proposed to tackle the challenging problem of automated posture evaluation during work activities.
Parsa et al.~\cite{Parsa2019} exploited spatial and temporal visual features from RGB-D frames using CNNs for predicting the ergonomic risk of object manipulation actions according to the REBA method.
Shafti et al.~\cite{Shafti} automatically extract 3D skeletal body information using RGB-D frames to continuously analyze the user’s posture and understand the safe range of arm motions during welding actions following the RULA posture monitoring method. 
Both methods in~\cite{SEO2021103725} and~\cite{YAN2017152} address the task of vision-based posture recognition on construction sites using the OWAS approach.
Recently, Parsa~et~al.~\cite{Parsa_2020_WACV,Parsa_2021_WACV} proposed a novel approach based on Graph-based CNNs and LSTMs to recognize object manipulation actions (lifting, moving boxes, etc.) and predict the REBA scores in videos.
Recently, a novel multi-stream deep network has been proposed \cite{Daras_2021_PETRA} to compute the REBA score based on 3D skeletal data sequences extracted during work activities. 
Our proposed framework introduces an efficient deep-learning method using Graph-based Neural Networks for recognizing the ergonomic postures using the 3D skeletal motion of workers during assembly activities in an industrial environment based on the widely used EAWS postural grid.

\subsection{Human actions monitoring}
\label{sec:sota_actions}

Understanding human actions in visual data is linked to advances in areas like human dynamics, domain adaptation, and semantic segmentation. Extensive research in this domain follows the latest deep learning methodologies~\cite{Herath2017, morshed_review}. Recently, much work has focused on human actions and human-object interactions (HOI) using deep neural networks~\cite{lecun2015deep,li2021symbiotic,9008780}. These approaches often model coarse geometric and appearance features using spatial regions-of-interest (ROIs)~\cite{gkioxari2018detecting} to classify actions in short video clips. More advanced methods represent the temporal and spatial structure of entities, such as 2D/3D skeletal models~\cite{cao2019openpose,Qammaz2020}, 2D hand/object masks~\cite{Baradel_2018_ECCV}, and 3D poses of hand(s)-object(s)~\cite{tekin2019h+,FirstPersonAction_CVPR2018}.

Spatiotemporal relationships in HOI are modeled using attention mechanisms~\cite{ma2018attend}, Graph Neural Networks (GNN)~\cite{li2021symbiotic}, CNNs~\cite{kim2017interpretable}, RNNs, LSTMs~\cite{zhang2017view, mahasseni2016_LSTM}, and Transformer models~\cite{zhang2021temporal,girdhar2019video, wu_memvit2022}. However, many methods treat HOI as non-composite, monolithic activities, limiting their ability to generalize across diverse actions. Recent research considers HOI as composite activities with complex spatiotemporal relationships~\cite{materzynska2020something}, integrating high-level semantics~\cite{bacharidis2021extracting}, logic rules~\cite{Xu_2019_CVPR}, and graph-based methods~\cite{papoutsakis2019unsupervised}. Novel representations encode human-object relationships over time using spatio-temporal graphs~\cite{ji2020action}, with some focusing on visual reasoning to uncover causal relationships~\cite{Baradel_2018_ECCV}. Our framework extends these methodologies to the structured context of industrial assembly lines, enhancing fine-grained activity recognition and ergonomic assessment.

\begin{table}[th]
\resizebox{\textwidth}{!}{
\begin{tabular}{|l|l|l|l|c|c|c|}
\hline
\multicolumn{1}{|c|}{\textbf{Dataset}}                           & Type of actions               & \multicolumn{1}{c|}{\textbf{Data}}                 & \textbf{Viewpoint (cameras)} & \textbf{Ergonomics} & \textbf{Videos} & \textbf{Duration (h)} \\ \hline
ATTACH~\cite{10160633}                     & assembly actions              & Video RGB-D, IR, 3D poses           & exocentric (3)         & -                   & 378             & 17                    \\ \hline
BRIO-TA~\cite{9897369}                     & assembly toy set              & Video (RGB)                         & exocentric (1 top-view)         & -                   & 75              & 2.9                   \\ \hline
IKEA ASM~\cite{Ben-Shabat_2021_WACV}     & assembly furniture              & Video RGB-D, 3D poses               & exocentric (3)         & -                   & 381             & 35.3                  \\ \hline
IndustReal~\cite{schoonbeek2024industreal} & maintenance, assembly toy set & Video RGB-D, Gaze                   & egocentric (1)        & -                   & 84              & 5.8                   \\ \hline
ENIGMA-51~\cite{ragusa2024enigma}          & repairing electrical board    & Video, audio instructions           & egocentric (1)        & -                   & 51              & 22                     \\ \hline
Assembly101~\cite{sener2022assembly101}    & assembly toy vehicles         & Video RGB-D, 3D hand poses          & exo-egocentric (8-4)  & -                   & 362             & 167                   \\ \hline
HA4M~\cite{cicirelli2022ha4m}              & assembly mechanical parts     & Video                               & exocentric (1)        & -                   & 217             & 5.9                   \\ \hline
sustAGE~\cite{technologies10020042}        & car door assembly             & Video RGB-D, heart rate             & exocentric (2)        & MURI                & 40              & 0.8                   \\ \hline
UW-IOM~\cite{Parsa2019}                    & manipulating boxes-rods       & Video RGB-D, 3D poses               & exocentric (1)        & REBA                & 20              & 0.63                  \\ \hline
AnDy~\cite{maurice2019human}               & screwing, manipulating loads  & Video RGB-D, motion capture         & exocentric (2)       & EAWS                & 390             & 5                     \\ \hline \hline
\textbf{CarDA}                             & \textbf{car door assembly}    & \textbf{Video RGB-D, motion capture} & exocentric (2)       & \textbf{EAWS}       & \textbf{25}     & \textbf{1.48}         \\ \hline
\end{tabular}
}
\caption{Overview of video datasets focusing on human activity monitoring in industrial or industrial-like environments.}
\label{tab:overview-datasets}
\end{table}

\subsection{Datasets}
\label{sec:sota_datasets}

We summarize available video datasets related to the visual understanding of human manufacturing activities in industrial environments, as shown in Table.~\ref{tab:overview-datasets}. Only, a few of them provide annotation data related to ergonomic analysis of body postures during work activities. The~UW-IOM dataset~\cite{Parsa2019} features limited object manipulation actions involving awkward poses and repetitions and frame-level annotations according to the REBA ergonomic risk index.
The sustAGE User Postures \& Actions Monitoring dataset~\cite{technologies10020042} includes time-synced video recordings from multiple static RGB-D cameras and heart rate data of real line workers acquired using wearable sensors recorded during manufacturing activities. Annotation data related to assembly actions and ergonomically sub-optimal postures based on the MURI postural grid for physical ergonomics assessment are provided.
Moreover, the AnDy multi-modal dataset~\cite{maurice2019human} captures human motions in industry-like activities (screwing and manipulating loads). Annotation data regards whole-body kinematics of $13$ participants recorded with optical motion capture, finger pressure force, videos, and postural events following the EAWS~\cite{schaub2013european}. 
The proposed CarDA dataset comprises videos (RGB-D) acquired using two stereo cameras, whole-body 3D kinematics acquired using a motion capture system, and annotations related to the assembly actions and the postures using the EAWS postural grid for analysis of physical ergonomics in a realistic industrial environment.
The IKEA ASM dataset~\cite{wang2020see} focuses on realistic chair assembly actions in videos, while the 
IndustReal \cite{schoonbeek2024industreal} and the ENIGMA-51~\cite{ragusa2024enigma} datasets comprise egocentric videos of maintenance or assembly activities on a construction-toy assembly set and repairing procedures on electrical boards, respectively. 
The Assembly101 dataset~\cite{sener2022assembly101} features exocentric and egocentric recordings of people assembling and disassembling 101 ``take-apart'' toy vehicles. 
The HA4M multi-modal dataset~\cite{cicirelli2022ha4m} demonstrates assembly activities for building an Epicyclic Gear Train.  
The ATTACH dataset~\cite{10160633} demonstrates simulated assembly actions in the context of a human-robot collaborative scenario. 
Finally, the BRIO-TA~\cite{9897369} dataset simulates operations in factory assembly consisting of scenarios for normal and anomalous work processes.

\section{CarDA - Car door Assembly Activities Dataset}
\label{sec:dataset}
The proposed multi-modal dataset for car door assembly activities noted as CarDA, comprises a set of time-synchronized multi-camera RGB-D videos and motion capture data acquired during car door assembly activities performed by real line workers in an experimental, real-world car manufacturing environment. The following sections provide a detailed description of the deployment environment, and the data acquisition and annotation procedures followed to compile the CarDA dataset. The dataset will be available freely available online~\footnote{~\url{https://zenodo.org/uploads/13370888}}.

\subsection{Deployment environment}
\label{sec:environ}

The use-case scenario concerns a real-world assembly line workplace in an automotive manufacturing industry, as the deployment environment. 
In this context, line workers simulate the real car door assembly workflow using the prompts, sequences, and tools under very similar ergonomic and environmental conditions as in existing factory shop floors. 
The assembly line consists of three workstations, overseen by a team leader. A conveyor belt with three virtually separated work areas that correspond to the three workstations, moves at a constant speed, supporting cart-mounted car doors and material storage. A worker is assigned to each workstation. All workers assemble car doors as the belt moves, with each station (WS10, WS20, and WS30). A worker completes a workstation-specific set of assembly actions, noted as a task cycle, lasting about 4 minutes before the cart proceeds to the next workstation for further assembly.
Upon the successful completion of the task cycle, the cart is left to travel to the virtually defined area of the subsequent workstation where another line worker will continue the assembly process during the new task cycle. 
Each task cycle lasts approximately 4 minutes and is continuously repeated during the worker’s shift. 

A camera system with three pairs of cameras is installed along the production line to monitor all workstations, capturing the assembly activities from both the inner and outer sides. 
Each RGB-D sensor, mounted on rods at a height of 1.85 meters, is oriented towards the center of a workstation and the conveyor belt for comprehensive visual monitoring.

\begin{figure}[t]
    \centering
    \includegraphics[width=0.75\linewidth]{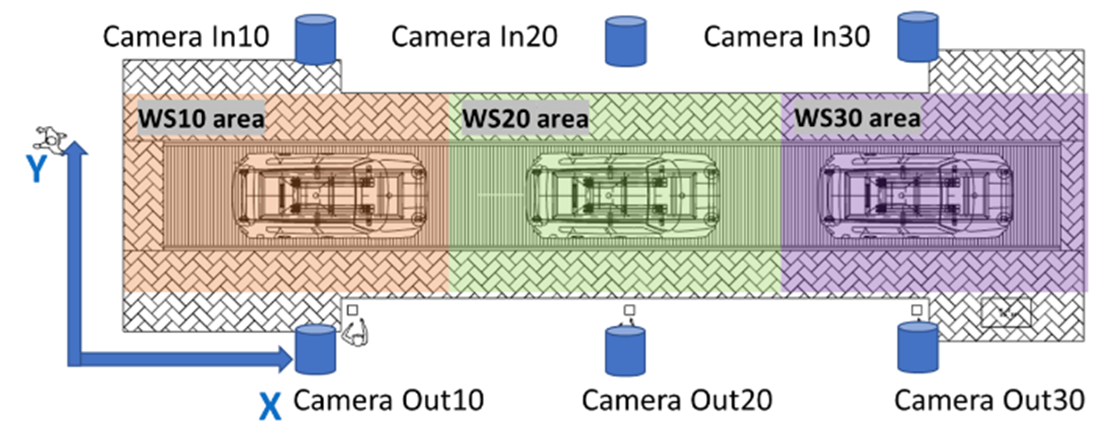}
    \caption{The shopfloor outline of the real manufacturing environment used to acquire the CarDA dataset. Three workstation areas (color-coded rectangles) are virtually defined on the conveyor belt area (assembly line). A pair of two stereo cameras (InXX, OutXX) (blue cylinders) are installed on both sides of each workstation WSXX.}
    \label{fig:camera_setup}
\end{figure}

\subsection{Car Door Assembly Activities}
The car door assembly dataset contains detailed timing data captured through multiple video recordings of the assembly process, as described above. This temporal data includes the start and end times of various activities, recorded in seconds. Each activity is part of a larger subgoal, which collectively describes the entire car door assembly process. Table~\ref{tab:door_assembly} presents indicative sub-goals for the first of the three workstations.
These subgoals span a period between 22 and 67 seconds each and occur sequentially during a complete assembly task cycle process. For the transformer model described in \ref{sec:actionMonitoring} we used these temporal annotations to extract the labels which correspond to the appropriate sub-goal. These labels were used to train the classifier.


\begin{table}[t]
    \centering
    \resizebox{\textwidth}{!}{%
        \begin{tabular}{|c|l|p{10cm}|}
            \hline
            \textbf{No} & \textbf{Sub-Goal} & \textbf{Description} \\
            \hline
            1 & Place Cart & Moving the cart at the assembly line to align the anterior wheels. \\
            \hline
            2 & Attach Door Frame & align and secure the door frame to the main body with bolts. \\
            \hline
            3 & Install Window Mechanisms & Fit motorized and manual window components into the door structure. \\
            \hline
            4 & Connect Wiring & Connecting the wiring harness to the door's electrical systems. \\
            \hline
            5 & Test Electrical Components & Verifying the functionality of electrical components (windows, locks). \\
            \hline
            6 & Install Interior Panels & Aligning and securing the inner door panel with screws and clips. \\
            \hline
            7 & Attach Exterior Components & Attach exterior components (handles and trim pieces) to the door. \\
            \hline
            8 & Quality Check & Perform visual inspection for defects, functional tests on moving parts. \\
            \hline
        \end{tabular}%
    }
    \caption{Indicative sub-goals for the assembly activities in WS10.}
    \label{tab:door_assembly}
\end{table}

\subsection{Data Acquisition}

Data acquisition involves low-cost, passive RGB-D camera sensors that are installed at stationary locations alongside the car door assembly line and a motion capture system for capturing time-synchronized sequences of images and motion capture data during car door assembly activities performed by real line workers.

More specifically, two stationary StereoLabs ZED2 stereo cameras were installed in each of the three workstations of the car door assembly line. The two stationary, workstation-specific cameras are located at bilateral positions on the two sides of the conveyor belt at the center of the area concerning that specific workstation. 
The pair of sensors were utilized to acquire stereo color and depth image sequences during car door task cycle executions.
Each recording comprises time-synchronized RGB (color) and depth image sequences captured throughout a task cycle execution at 30 frames per second (fps).
At the same time, the line worker used a wearable XSens MVN Link suit during work activities to acquire time-synced 3D motion capture data at 60 fps.
The motion capture system and the pair of RGB-D cameras are calibrated according to a common (global) coordinate reference system set near the start of the assembly line.
A complete task cycle execution has an average duration of 4 minutes and comprises a fixed sequence of car door assembly actions (also noted as assembly subgoals). 
Two RGB videos, a sequence of depth images, and a BHV file with motion capture data are recorded per task cycle execution.
Overall, $25$ recordings are available for an equal number of car door task cycles, noted as samples: 8 recordings concern WS10, 10 for WS20, and 7 for WS30. 
We note that all CarDA samples contain videos, while 12 of the samples correspond to both video and motion capture data (5 for WS10, 4 for WS20, and 3 for WS30).

\begin{figure}[t]
    \centering
    \includegraphics[width=0.31\textwidth]{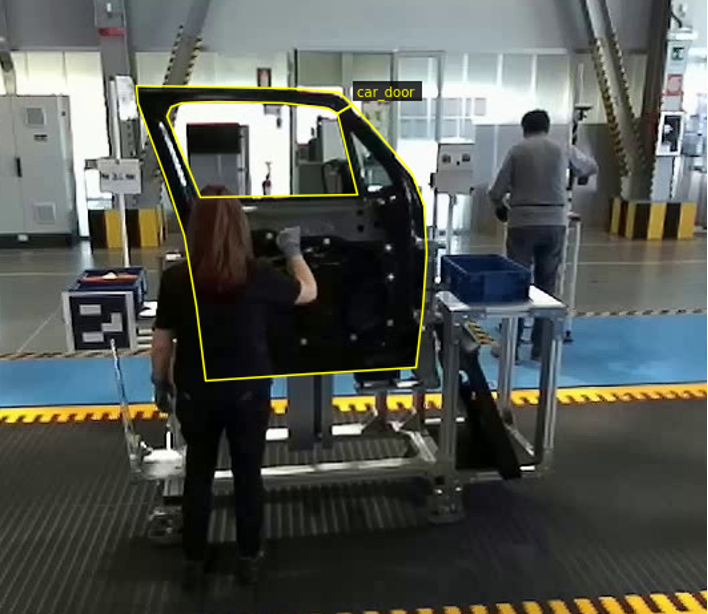}
    \includegraphics[width=0.31\textwidth]{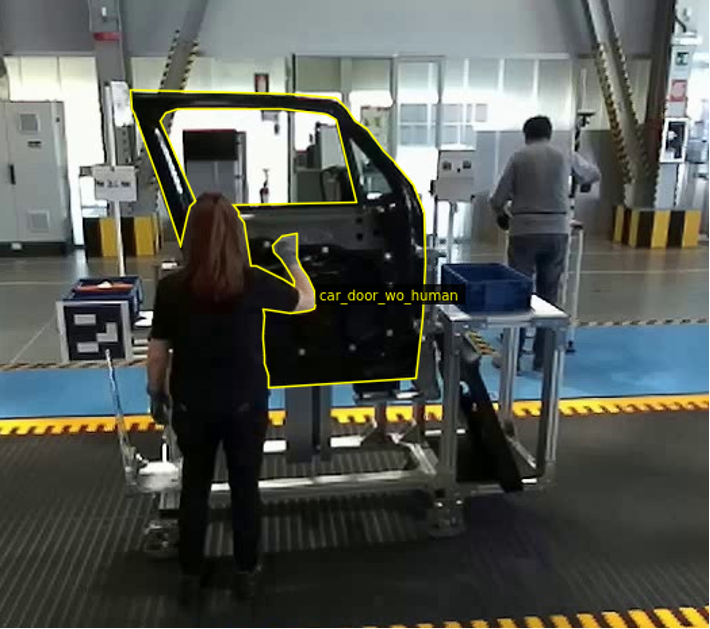}
    \includegraphics[width=0.31\textwidth]{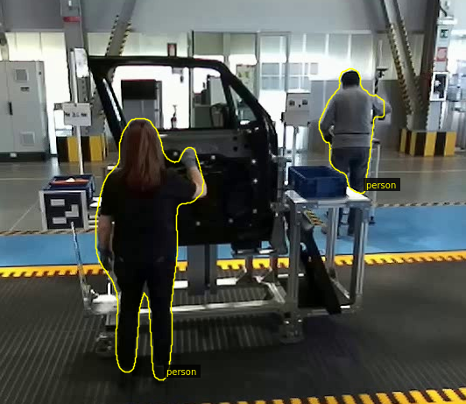}
    \caption[Annotations of the car door and humans regions]{Annotations of the car door and humans area (marked in yellow). Annotations of the car door with (left) and without (middle) the overlapping human area and the extracted 2D segmentation masks of the humans (right).}
    \label{fig:hbu_annotation}
\end{figure}

\subsection{Data Annotation}

\paragraph{\textbf{Ground truth data on human localization and 3D pose estimation:}}
As described above, ground truth data for 3D human motion is acquired using the wearable XSens MVN Link suit. 
The motion capture system obtains the 3D coordinates and orientations of 24 skeletal body joints throughout a recording in a BVH format, that corresponds to a task cycle execution in our case.

\paragraph{\textbf{Ground truth data on basic postures of the EAWS ergonomic screening tool:}}

Two experts in manufacturing and ergonomics performed manual annotations related to the EAWS screening tool.
Overall, the CarDA dataset contains videos of 25 task cycles for all workstations for which 412 instances concerning the seven target classes of EAWS basic postures are obtained. The average duration of annotated postures is approximately 10 seconds.
For each activity subgoal, a single occurrence, if any, and its total duration for any of the posture classes that have been observed is annotated, as shown in Tab.~\ref{tab:annotations}.
Information for the id-name of each subgoal, and temporal segmentation (start, and end video timestamps, and the duration of each subgoal in seconds) is provided. The total duration of each type of EAWS-based basic posture, if any occurred, during each subgoal is also reported. Postures of duration lower than 4 seconds are not considered valid for postural evaluation.


\begin{table}[t]
\resizebox{\textwidth}{!}{%
\begin{tabular}{|l|cc|c|c|c|c|c|c|}
\hline
\begin{tabular}[c]{@{}l@{}}EAWS evaluation\\ (Video: WS10\_HD720)\end{tabular} & \multicolumn{2}{l|}{Temporal segmentation}       & \multicolumn{1}{l|}{3. Bent forward} & \multicolumn{1}{l|}{\begin{tabular}[c]{@{}l@{}}4. Strongly \\ bend forward\end{tabular}} & \multicolumn{1}{l|}{\begin{tabular}[c]{@{}l@{}}5. Elbow at/above \\ shoulder level\end{tabular}} & \multicolumn{1}{l|}{\begin{tabular}[c]{@{}l@{}}6. Hands above \\ head level\end{tabular}} & \multicolumn{1}{l|}{\begin{tabular}[c]{@{}l@{}}Trunk\\ Rotation\end{tabular}} & \multicolumn{1}{l|}{\begin{tabular}[c]{@{}l@{}}Lateral\\ Bending\end{tabular}} \\ \hline
Subgoal                                                                        & \multicolumn{1}{c|}{Start {[}s{]}} & End {[}s{]} & Duration (s)                         & Duration (s)                                                                             & Duration (s)                                                                                     & Duration (s)                                                                              & Duration (s)                                                                  & Duration (s)                                                                   \\ \hline
1. Adjust WS                                                                   & 11                                 & 52          & 16                                   &                                                                                          &                                                                                                  &                                                                                           &                                                                               &                                                                                \\ \cline{1-1}
2. Fixate rearview mirror                                                      & 52                                 & 102         & 15                                   &                                                                                          & 19                                                                                               &                                                                                           & 10                                                                            & 19                                                                             \\ \cline{1-1}
3. Fixate tie rod                                                              & 102                                & 151         & 21                                   & 1                                                                                        &                                                                                                  &                                                                                           &                                                                               & 21                                                                             \\ \cline{1-1}
4. Fixate plugs                                                                & 151                                & 180         & 6                                    &                                                                                          &                                                                                                  &                                                                                           &                                                                               &                                                                                \\ \cline{1-1}
5. Fixate carrier                                                              & 180                                & 269         & 5                                    &                                                                                          &                                                                                                  &                                                                                           &                                                                               & 10                                                                             \\ \hline
\end{tabular}%
}
\caption{Sample annotation data related to EAWS-based ergonomic postures and work/assembly activities (subgoals) for a complete task cycle in WS10.}
\label{tab:annotations}
\end{table}

\paragraph{\textbf{Ground truth related to temporal segmentation and classification of car door assembly actions (subgoals) during task cycle executions:}}
Manual annotations related to the temporal segmentation and classification of car door assembly actions were performed by personnel working directly on the assembly line for the CarDA dataset, as shown in Tab.~\ref{tab:annotations}.
It encompasses 25 task cycles (samples), where each comprises on average 5.4 subgoals (minimum of 4 and maximum of 8 subgoals) and has a mean duration of 216 seconds, ranging from 137 to 271 seconds. 
In total, the set of 25 task cycles comprised 135 different subgoals that were manually for the start and end timestamps (in seconds of video time).
Each subgoal has an average duration of 42 seconds, ranging from 22 to 67 seconds.

\section{Human behavior understanding framework}
In the following sections, the proposed framework for human behavior understanding in the assembly line work environment is presented (Fig.~\ref{fig:hbu_workflow}). It relies on visual data that capture stereo RGB (color) and depth images of human workers during work activities on the real shop floor and comprises a set of components and methods for a) detecting the location and the 3D body pose of the human workers in real-time; b) analyzing the spatiotemporal relationships of body joints for monitoring workers’ postures in terms of physical ergonomics and c) estimate the work progress during an ongoing assembly task cycle execution. 
The framework also caters to the estimation of the location of each cart door during assembly activities which will aid in real-time monitoring of worker actions and the task cycle progress estimation.

\subsection{Human and car door detection}
\label{sec:human_cardoor}

For the estimation of the human pose, we exploit two state-of-the-art methods to estimate the 2D and subsequently the 3D body articulated (skeleton-based) configuration based on a known tree-based skeletal model. First, the OpenPose~\cite{cao2019openpose} deep learning-based method is employed to detect the 2D location of each human in a single image and estimate the 2D body joint coordinates of 25 body joints (Model25). 
Then, the set of 2D poses is used as input to the MocapNet2~\cite{Qammaz2020} method, an end-to-end approach that relies on ensembles of Deep Neural Networks to regress the 3D coordinates and angles of $15$ human body joints per frame as in~\cite{papoutsakis2022}. 
The method is further extended to process visual data captured synchronously by a pair of cameras installed on both sides of each workstation of the assembly line (cf.~\ref{sec:environ}) to improve the reliability of the recovered 3D human body pose. Given the confidence values of the 2D and 3D estimations for the body joint positions, we opt for the camera view that provides valid detections (more than  $50\%$ of body joints provide a confidence score higher than $50\%$). 
We utilize the pre-trained Mocapnet2 method without any additional training based on the proposed CarDA dataset. 

Beyond the human pose, the car door is an instrumental element in the analysis of human assembly activities. The estimated car door pose (position and orientation) is used in conjunction with the skeletal joints in the human actions monitoring component (sec.~\ref{sec:human_actions}). To estimate the car door pose, the car door area is visually segmented 
and the obtained fine object mask is subsequently exploited to estimate the car door centroid, orientation, and bounding box both in camera-centric coordinates and global coordinates. For the car door segmentation, the Mask R-CNN deep learning model~\cite{maskRCNN} was exploited as part of the Detectron2\footnote{https://ai.meta.com/tools/detectron2/} framework. 
Initially, we used the Mask R-CNN R50 FPN model from the set of pre-trained CNN models from the model zoo, trained on the COCO dataset, to assess the extraction of different class types, including the segmentation of humans (i.e., class \emph{person}). 
Since the car door is a special class of object not part of available pre-trained models, the car door area had to be annotated in sample images from a series of videos from the assembly line and further 
finetune the model for the newly introduced car door class together with the 2D segmentation masks of humans (i.e., the class for \emph{person}). 

The different experiments for finetuning the model either exploiting only the \emph{car door} or both the \emph{person} and \emph{car door} classes and validating the results on image data from available video sequences are presented in sec.~\ref{sec:exp_car}.

Once the 2D segmentation mask of the car door is detected, we utilize the depth information and the camera extrinsic parameters to estimate the car door pose. 
By transforming the mask's extremities to the camera and global coordinate system, we deduce the door's orientation and centroid. 
Furthermore, the 2D human masks extracted with Mask R-CNN are also used to extract the skeletal data in each dilated human mask area and select the optimal 3D human pose based on the 3D distance to the cart location. 
This significantly improves the results in the case of multiple humans present in the scene as in these cases the assumption to select the 2D pose with the highest confidence score as the optimal candidate and active worker for 3D pose estimation would be less robust.

\subsection{Analysis of ergonomic work postures in videos}
\label{sec:ergonomics}
We focus on seven types of dynamic body postures based on the basic postures of the EAWS screening tool~\cite{maurice2019human,schaub2012ergonomic,schaub2013european}. 
The EAWS tool is commonly employed in manufacturing to assess workers' physical workload and ergonomics.
Each basic posture type is treated as a time-varying event, consisting of a sequence of body configurations with a minimum duration of 4 seconds. The EAWS tool uses a scoring/labeling scheme based on the occurrences and total duration of each posture type per minute throughout a work activity. 
Some of the basic postures concern similar types of body configurations with increased levels of body strain. 
The selected posture types are the following  
(by keeping the original numbering/naming): 
1. Standing \& walking in alteration, 3. Bend forward, 4. Strongly Bend forward, 5. Upright with elbows at or above shoulders, 6. Upright with elbows above the head, and finally the "Trunk
Rotation" and the "Lateral Trunk Bending" types from the Asymmetric postures.
Static postures of duration at least 4 seconds are identified as valid, according to the EAWS instructions. 


We formulate the evaluation of ergonomic work postures during manufacturing activities as a vision-based multi-class action detection and classification problem. 
To this end, given a video demonstrating a subgoal during a car door task cycle for a specific workstation, we extract the sequence of 3D body poses using the MocapNet2 method for each frame. 
We adapt and re-trained the previously proposed approach in ~\cite{technologies10020042} for the detection and classification of EAWS-based postures throughout a video.
The 3D skeletal sequences are used to re-train a spatiotemporal Graph Convolutional Network model (ST-GCN)~\cite{stgcn2018aaai}, which learns to encode the input representations of the target working postures into a shared embedding/feature space. Finally, we compute the pairwise temporal alignment cost between the embeddings of an unlabeled 3D skeletal sequence and those of a 3D skeletal sequence of a known class type using the soft Dynamic Time Warping approach (softDTW)~\cite{soft_dtw_blondel}. 
This pairwise cost serves as a similarity measure for classifying the unlabeled sequence among the target types of working postures. 
We employ a sliding window scheme of 8 seconds, with an overlap of 4 seconds, to detect and classify the occurrences of any ergonomic posture classes using our classifier.


\subsection{Human actions monitoring }
\label{sec:actionMonitoring}

To monitor and interpret actions in the car door assembly line, a transformer model was chosen for its ability to capture long-range dependencies in sequence-to-sequence tasks~\cite{zhang2023longrange}. This model processes input data, including the car door's location, the human operator's position, and skeletal joint measurements, all mapped in world coordinates to maintain consistency~\cite{vaswani2017attention, devlin2019bert}.  The car door's position is derived from the center of mass of its segmentation mask, identified using the Mask R-CNN model (cf. Section~\ref{sec:human_cardoor}).Human skeletal data is transformed into a body coordinate system, transposing movements relative to the base spine joint, which is set as the origin (0,0). All other joints are transposed relative to this base joint, focusing on relative movements rather than absolute positions. This transformation normalizes the data and helps the model learn movement patterns effectively. .

The model handles 10 non-overlapping sets of measurements, embedding spatial and kinematic data. Using transformer encoder layers with multi-head self-attention, the model condenses the output into a vector, which is classified through a fully connected and softmax layer. The model is trained using cross-entropy loss and the Adam optimizer, with a 70-10-20 train-validation-test split, to infer the subgoal of the assembly task based on these measurements.

\label{sec:human_actions}

\begin{figure}[t]
\centering
\begin{subfigure}{.5\textwidth}
  \centering
  \includegraphics[width=.85\linewidth]{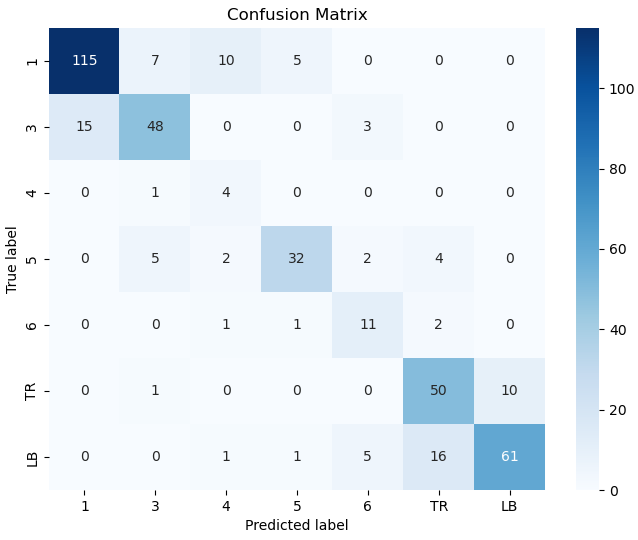}
  \caption{EAWS-based posture classification}
  \label{fig:sub1}
\end{subfigure}%
\begin{subfigure}{.5\textwidth}
  \centering
  \includegraphics[width=.85\linewidth]{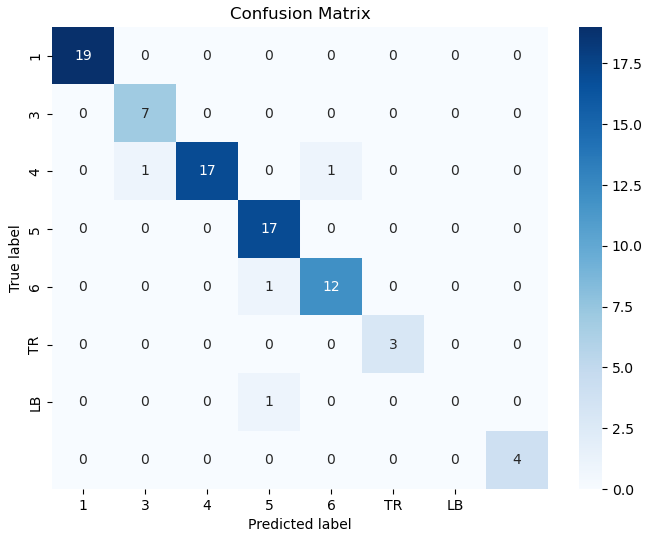}
  \caption{Assembly progress monitoring}
  \label{fig:sub2}
\end{subfigure}
\caption{(a) Experimental results are illustrated as a confusion matrix for (a) the vision-based classification of EAWS-based ergonomic postures (TR: trunk rotation, LB: lateral bend) as shown in Sec.~\ref{sec:ergonomics}, (b) the progress monitoring of car door assembly activities during task cycles using the proposed Car Door Assembly dataset.}
\label{fig:results_confusion_m}
\end{figure}

\section{Experiments}

\begin{table}[t]
\resizebox{\textwidth}{!}{%
\begin{tabular}{|l|c|c|c|c|c|c|c|c||c|}
\hline
Metrics/Joints & Neck & Trunk & Knees & Shoulders & Elbows & Hands & Hips  & Toes  & \textbf{Mean}  \\ \hline \hline
MJPE           & 98.4 & 115.4 & 156.4 & 102.7     & 157.5  & 171.8 & 127.8 & 175.6 & \textbf{138.2} \\ \hline
PCK@150        & 81.3 & 82.9  & 56.4  & 81.1      & 58.6   & 49.3  & 75.6  & 65.2  & \textbf{68.8}  \\ \hline \hline
\end{tabular}%
}
\caption{Experimental results for 3D pose estimation using the MocapNet2 method~\cite{Qammaz2023b}. MPJPE (Mean Per Joint Position Error).  PCK@150 (Percentage of Correct Key-points metric between 3D poses using the commonly used threshold of 150mm). The average scores for pairs of symmetric body joints are reported, e.g. hands.
 }
 \label{tab:pose_estimation}
\end{table}

\begin{figure}[t]
\centering
\begin{subfigure}{.54\textwidth}
  \centering
  \includegraphics[width=\linewidth]{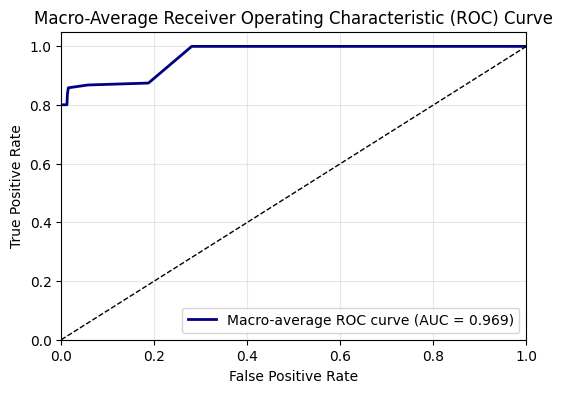}
  \caption{ROC Curve}
  \label{fig:task_analysis_sub1}
\end{subfigure}%
\begin{subfigure}{.4\textwidth}
  \centering
  \includegraphics[width=\linewidth]{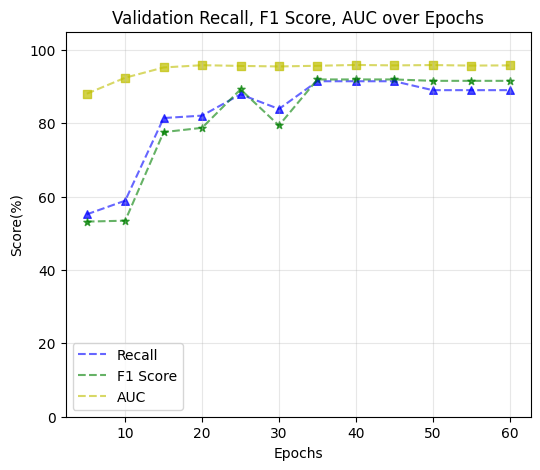}
  \caption{Validation Recall, F1 Score, AUC over Epochs}
  \label{fig:task_analysis_sub2}
\end{subfigure}
\caption{ Performance evaluation of the human action monitoring (a) Macro-Average Receiver Operating Characteristic curve, (b) Recall, F1, AUC scores (validation set).}
\label{fig:task_analysis}
\end{figure}

\subsection{Car Door Detection}
\label{sec:exp_car}
We used $257$ frames from ~\cite{technologies10020042} for annotation for the \emph{car door} class and derived two sets of annotation data.  
The first set comprised masks of the car door with overlapping human areas in the delineated car door area, while in the second set, mask parts attributed to humans were excluded from the estimated car door area/mask (Figure~\ref{fig:hbu_annotation}). 
Initially, we trained the model individually with the two sets of car door classes and without the 2D segmentation masks for humans. Overall, the results with the first set exhibited sufficient accuracy, with 85.07\% and 80.86\%  average precision for the bounding boxes and the segmentation masks, respectively. 
With the second set, the average precision on segmentation masks was lower (66.98\%) due to the intricate shape of the car door annotations. Overall, the car door segmentation masks were correctly estimated in cases where humans were not present.
We have further fine-tuned the model using annotation data of both sets containing the 2D mask of the car door and those of the humans. The results did not exhibit significant differences in the way the model dealt with the prediction of the two classes, as
both the car doors and humans were correctly detected with a confidence score above 95\%, for both classes. The average precision with the COCOEvaluator is 88.20\% on bounding boxes and 79.63\% on segmentation masks.

\subsection{Vision-based Human Detection}

We evaluate the performance of the MocapNet2~\cite{Qammaz2020,Qammaz2023b} method for 3D human body pose estimation using ground truth motion capture data and visual information from the CarDA dataset. A subset of $18$ samples of the proposed dataset is used for this evaluation. 

Each sample comprises time-synchronized video acquired by one camera and ground truth 3D motion capture data acquired using the Xsens MVN Link motion capture system during assembly activities.
For each image sequence, the 3D human poses are computed (cf. Section~\ref{sec:human_cardoor}). 
To compare the estimated 3d pose to the ground truth data, we consider the 3D body center (Torso) position as the origin of the coordinate system for both sequences of 3D skeleton-based motion data. The pre-trained MocapNet2 model was deployed, with no fine-tuning on the ground truth motion capture data. Moreover, a generic 3D human body model is used to infer the 3D human pose per frame. 

Firstly, the Mean Per Joint Position Error (MPJPE) is used to assess the average Euclidean distance between the 2D or 3D positions of predicted joints of the estimated skeleton-based human pose and the ground truth joints in a given dataset. 
Moreover, the Percentage of Correct Keypoints (PCK) metric is used. The PCK considers correctly estimated detected joints as the ones that have a distance below a certain threshold compared to the true location. 
We use a fixed threshold of $150$ mm that is commonly used in pose estimation benchmarks.
Table.~\ref{tab:pose_estimation} provides the experimental results, where the average MPJPE over all joints and all samples of the CarDA subset is below 140 mm and $68.8\%$ of the 3D joints are considered successfully detected throughout all data sequences based on the PCK@150 metric.


\subsection{Vision-based analysis of Ergonomic Work Postures} 
The experimental evaluation of vision-based physical ergonomics is posed as an action classification task given the 25 samples of the CarDA dataset and available EAWS-based annotations.
We estimate the performance of the proposed approach based on the accuracy and precision scores, as shown in Tab.~\ref{tab:eaws_table_scores} and Tab.~\ref{fig:sub1}.
Overall, the proposed approach can identify the correct class of postures performed by line workers with $76.7\%$ accuracy and $68.27\%$ mAP scores.
Given the challenging conditions captured by the CarDA dataset in a real environment of simulated manufacturing activities, the obtained results indicate a promising potential for the proposed approach in real-world applications. 
These findings demonstrate the effectiveness of vision-based physical ergonomics in accurately classifying worker postures, even under complex and dynamic conditions, and severe and long-term body occlusions. 



\begin{table}[t]
\resizebox{\textwidth}{!}{%
\scalebox{0.3}{
\begin{tabular}{|l|c|c|c|c|c|c|c||c|}
\hline
\begin{tabular}[c]{@{}l@{}}Postures \& \\ evaluation metrics\end{tabular} & 1      & 3      & 4      & 5      & 6      & \begin{tabular}[c]{@{}c@{}}Trunk \\ rotation\end{tabular} & \begin{tabular}[c]{@{}c@{}}Lateral \\ Bending\end{tabular} & \textbf{Mean}   \\ \hline \hline
Accuracy                                                             & 0.8394 & 0.7273 & 0.8000 & 0.7273 & 0.7333 & 0.8197                                                    & 0.7262                                                     & \textbf{0.7670} \\ \hline
Precision                                                            & 0.8846 & 0.7742 & 0.2222 & 0.8205 & 0.5238 & 0.6944                                                    & 0.8592                                                     & \textbf{0.6827} \\ \hline \hline
\end{tabular}%
}
}
\caption{Experimental results for the vision-based classification of EAWS-based ergonomic work postures (as described in Sec.~\ref{sec:ergonomics}) using the Car Door Assembly dataset. The mean accuracy and 
 mean average precision (mAP) are reported (last column).}
\label{tab:eaws_table_scores}
\end{table}


\subsection{Vision-Based Progress Monitoring of Assembly Activities}




The performance of the transformer model for human actions monitoring in the car door assembly line demonstrates strong results (Figure~\ref{fig:task_analysis}). The first graph shows the validation recall, F1 score, and AUC over 60 epochs, indicating significant improvements early in the training process, with the metrics stabilizing at high values, around 90-100\%. The macro-average ROC curve reflects an AUC of 0.969, signifying excellent discriminative capability. The confusion matrix further supports these findings, showcasing precise classification across different subgoals, with most true labels accurately predicted, and minimal misclassifications, particularly for subgoals with fewer samples. These results confirm the model's robustness and effectiveness in real-time monitoring applications.


\section{Conclusion}

In this paper, we have introduced a comprehensive vision-based framework for real-time monitoring of human behavior in industrial settings, specifically focusing on the manufacturing sector and the car door assembly process. We introduced the CarDA dataset, a new dataset that comprises time-synchronized multi-camera video with 3D motion capture data of trained line workers during assembly activities in a real-world manufacturing environment. A rich set of annotations, related to the assembly activities, motion capture data, and information on EAWS-based physical ergonomics is provided to support the multifaceted human behavior. The utilization of the MocapNet2 method allowed for robust 3D pose estimations in real-time, which are crucial for accurate ergonomic assessments. Moreover, a transformer-based approach in presented to monitor the work progress during assembly activities using visual information on the human and car door motion in the work environment. Yet another deep learning-based approach for the classification the EAWS work postures using sequences of 3d skeletal poses is presented. Despite the challenging conditions presented in the CarDA dataset, our approach achieved commendable performance, highlighting its robustness and reliability. Future work will focus on enhancing the model's performance by incorporating additional data and refining the pose estimation techniques. Furthermore, we aim to explore the applicability of our framework to other industrial processes and ergonomic evaluation protocols, thereby broadening its impact on occupational safety and health.

\section*{Acknowledgements}
This work was funded by the European Union’s Horizon 2020 programme FELICE (GA No 101017151) European Union’s Horizon Europe programme SOPRANO (GA No 101120990). Special thanks to Stellantis—Centro Ricerche FIAT (CRF)/ SPW Research \& Innovation department in Melfi, Italy for supporting the dataset collection.

%
%
\bibliographystyle{splncs04}
\bibliography{main}
\end{document}